# Expressing Relational and Temporal Knowledge in Visual Probabilistic Networks

L. Enrique Sucar, Duncan F. Gillies
Department of Computing
Imperial College of Science, Technology and Medicine
180 Queen's Gate
London SW7 2BZ
England

## Abstract

Bayesian networks have been used extensively in diagnostic tasks such as medicine, where they represent the dependency relations between a set of symptoms and a set of diseases. A criticism of this type of knowledge representation is that it is restricted to this kind of task, and that it cannot cope with the knowledge required in other artificial intelligence applications. For example, in computer vision, we require the ability to model complex knowledge, including *temporal* and *relational* factors. In this paper we extend Bayesian networks to model relational and temporal knowledge for high-level vision. These extended networks have a simple structure which permits us to propagate probability efficiently. We have applied them to the domain of endoscopy, illustrating how the general modelling principles can be used in specific cases.

## 1 INTRODUCTION

Probabilistic networks (causal networks, Bayesian networks [Pearl 86]) have been used mainly in diagnostic tasks, for example in medicine [Lauritzen 88], where they represent the dependency relations between a set of symptoms and a set of diseases. In fact, a criticism of this type of knowledge representation is that it is restricted to this kind of tasks [Fox 90, Quinlan 83], and that it is too restrictive for the knowledge required in other artificial intelligence applications. Other formalisms, such as predicate calculus, have a rich expressive power for modelling our world knowledge. Temporal logics have been developed for reasoning in dynamic environments, and semantic networks and frame systems help to express relations and structures in the knowledge-base.

Reasoning in computer vision requires the ability to model complex knowledge, including *temporal* and *relational*. Recognition of objects from their parts requires expression of the toplogical relations between them (above, below, adjacent, ...). For example, an "arc" is described by two *non-adjacent* bricks that *sustain* another brick [Winston 75]. Temporal reasoning is also important, mainly in navigation, where information from a sequence of images is required. Features from previous images can be useful for interpreting the current image, and some objects can only be recognised from a sequence of images. For example, in endoscopy (see section 4) to detect that there is a *dirt spot* on the lens we need to identify an "static" region in a sequence of images as the endoscope moves.

In this paper we develop a methodology to model relational and temporal knowledge using a probabilistic network representation, and illustrate its application in an expert system for endoscopy.

## 2 EXPRESSING RELATIONAL KNOWLEDGE

Visual understanding implies the need to have an internal representation of our knowledge about the physical world. This allows us to recognize the objects of interest in a certain domain, and also gives feedback to the low-level vision processes for extracting the relevant features from the image(s). A visual knowledge-base (KB) relates the features to the objects, through intermediate regions, surfaces and sub-objects, depending on the complexity of the objects in the domain. Thus, we can think of a visual KB as a series of hierarchical relationships, starting from the low-level evidence features, to regions, surfaces, sub-objects, and objects. Several knowledge representations have been proposed for modelling a visual KB (see Rao [88] for a review). We proposed a representation based on a probabilistic network [Sucar 91] in which the network as divided into a series of layers. The nodes of the lower layer correspond to the feature variables and the nodes in the upper layer to the object variables. The intermediate layers have nodes for other visual entities, such as parts of an object or image regions. The arrows point from nodes in the upper layers toward nodes in the lower layers, expressing a causal relationship. Probabilistic networks have also been used to represent geometrical models in model-based vision [Binford 89].

In general probabilistic models express only unary relationships between objects and features. That is, the presence of a certain object in the world will *cause* the appearance of some specific features in the image, which are usually assumed independent. So there is a unary relationship between an object $O$ and a feature $F$. This is represented graphically by the probabilistic network in figure 1(a). Multiple objects in the world are easier to model as the composition of several sub-objects or parts, subdivided recursively until at the lowest level there are simple objects that will generally correspond to a region in the image. An example of this type of representation are the 3-D models proposed by Marr [82] in which an object is described by a number of cylindrical parts and their relationships. In this case the recognition of an object is not only dependant on the presence of its subparts, but it is also dependant on the physical relation between them. So we need to represent this dependency of the object on the relation between its components in the network.

The physical relation between the components can be directly determined from the image, for example the distance between two regions or their topological relation (adjacent, above, surrounding, etc.), so we can say that there is a *deterministic link* from the components to their relation. The presence of this image relation can be caused by the object of interest, but it could also be accidental or illusory depending on the view point, so we consider that there is some *probabilistic link* from the object to the relation. A deterministic link represents a deterministic relation between the variables, in contrast with the usual probabilistic interpretation in probabilistic networks. This means that the value of the variable at the end is a function of the values of its causes. For the case of one object and a binary relation between two features, the corresponding network is shown in figure 1(b). A similar extension of probabilistic networks to include deterministic variables has been done before in what are called *influence diagrams* [Shachter 87]. We will represent a probabilistic link by an arrow and we will represent the functional or deterministic links by a dotted arrow as shown in figure 1(c).

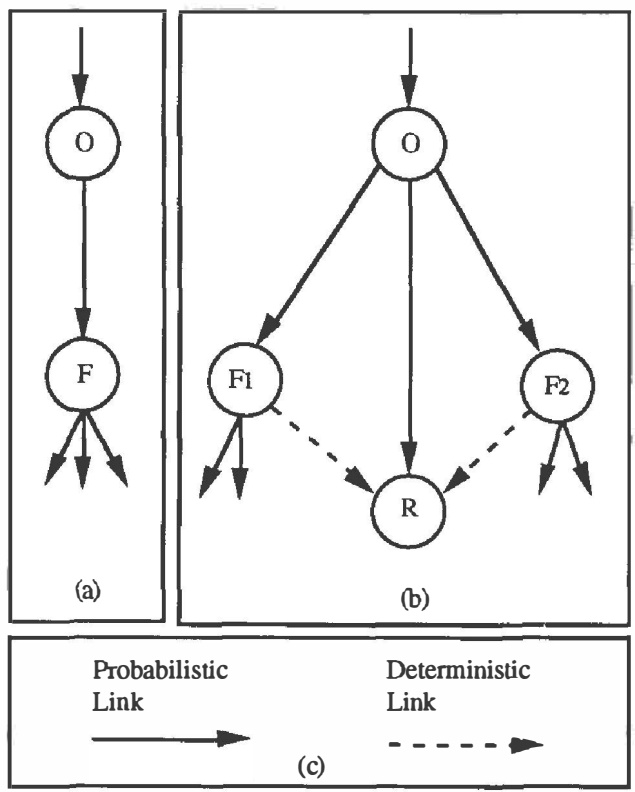

Figure 1: Expressing relational knowledge. The network in (a) represents a unary relation between $O$ and $F$, and the network in (b) a binary relation between $O$ and $F_1$, $F_2$ which is expressed through the relational variable $R$. The deterministic links between $F_1$, $F_2$ and $R$ indicate that the value of $R$ is a deterministic function of $F_1$ and $F_2$.

We can extend the representation to include more variables, and in general a relation for $n$ variables can be expressed by an analogous network as depicted in figure 2(a). Assuming that the relation can be uniquely determined from the related variables, the node $R$, shown as a shaded node in the network, becomes a deterministic variable similar to the input variables. Thus, if we eliminate the functional links from the network, we are left with the probabilistic network shown in figure 2(b). This network has a tree structure so we could apply efficient algorithms developed for probability propagation in trees [Pearl 86].

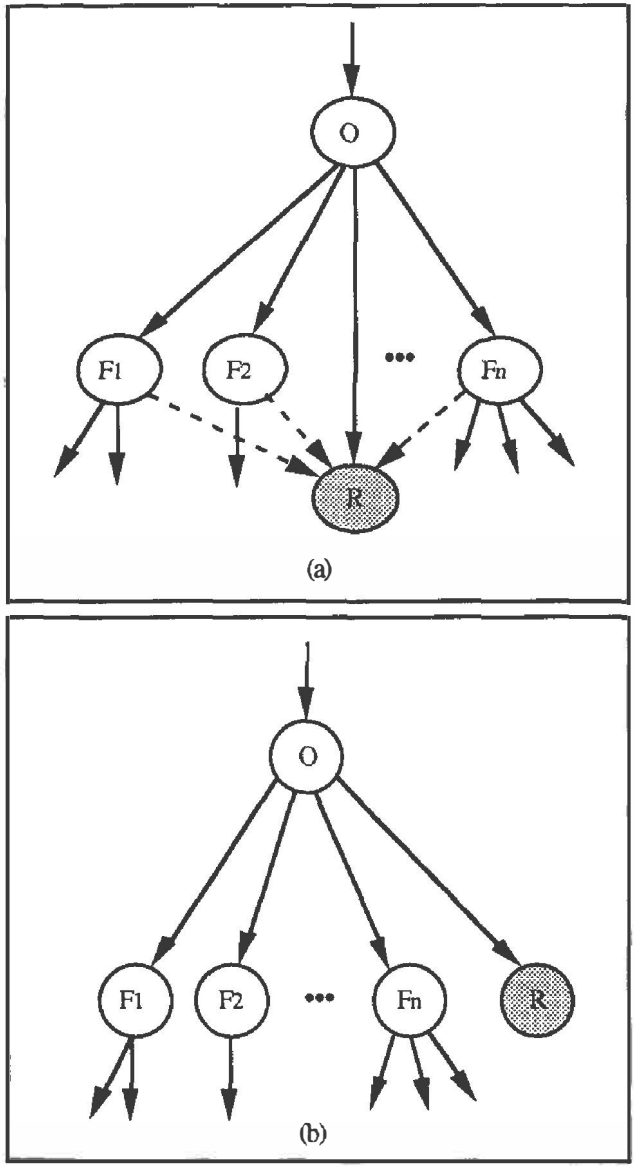

such as in endoscope navigation, we are required to incorporate in our knowledge-base temporal information which can be used efficiently as an aid for recognition. We focus on the navigation aspect, considering a series of images that are similar, with an incremental difference between one image and the next due to the movement of the camera or the environment; and we consider two cases:

(a) *Semi-static recognition*, in which an object can be identified from a single image but observations from previous images can be useful as another clue for recognition.

(b) *Dynamic recognition*, in which a series of images are necessary to identify an object, which otherwise will be impossible or very difficult to recognize.

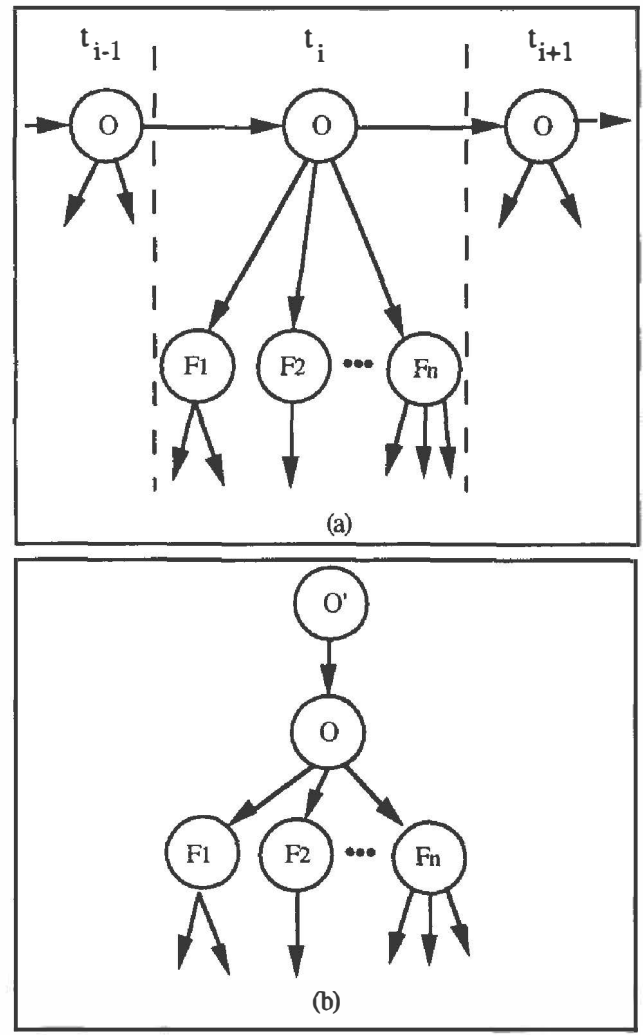

Figure 2: Probabilistic network for an $n$-variable relation. The network in (a) expresses the functional and probabilistic dependencies, which is transformed into the equivalent probabilistic network depicted in (b). The relation variable $R$ is treated as an instantiated variable drawn as a shaded circle.

## 3 EXPRESSING TEMPORAL KNOWLEDGE

We live in a dynamic environment in which the information we obtain from our visual sensors changes constantly as a function of our movement or the movement of the surrounding objects. We can think of this dynamic information as a series of images which together provide more information than a single static image. To be able to operate in a dynamic environment,

Figure 3: Semi-static recognition. The network in (a) represents the causal relations between an object in several images. This is simplified to the network in (b) for recognition at time $t_i$.

In semi-static recognition we consider an object $O$ which can be identified by a number of image features $F_1, ..., F_N$ obtained from an image at time $t_i$. Assuming that the time difference $\Delta t$ between images is small, the same type of object will be present at the same location in previous and successive images. We can represent this as object $O$ appearing at time $t_{i-1}$ *causes* the presence of the same object at time $t_i$, with an analogous relation between $t_i$ and $t_{i+1}$. This can be represented graphically by the network in figure 3(a), where the conditional probabilities for the links connecting the matching objects at different times will be *high* for objects of the same type and *low* for objects of different types. For evaluating the network at time $t_i$, we can consider only the node for the object at time $t_{i-1}$ obtaining the structure shown in figure 3(b). This simplification is based on the assumption that knowing the identity of the object at $t_{i-1}$ makes the previous observations irrelevant for $O$ at $t_i$. Thus, the posterior probabilities for an object in the previous image will influence the prior probability of the corresponding object in the current image. We will continue to have a tree structure although the object of interest is not longer at the root of this tree.

In dynamic recognition, an object is defined by the presence of other objects or certain image features across a series of successive images. We can think of this case as analogous to the hierarchical description of an object in terms of parts, regions, etc., but in which the sub-objects are in different images instead of the same one. Then we can represent a dynamic object with a probabilistic tree, where the presence of object $O$ is dependent on the observation of objects $S_1, ..., S_N$ in different successive images, as it is shown in figure 4(a). The physical relation, such as the distance, between these objects could be also important and we can extend the representation by including this relational information in the way described in the previous section (figure 4(b)).

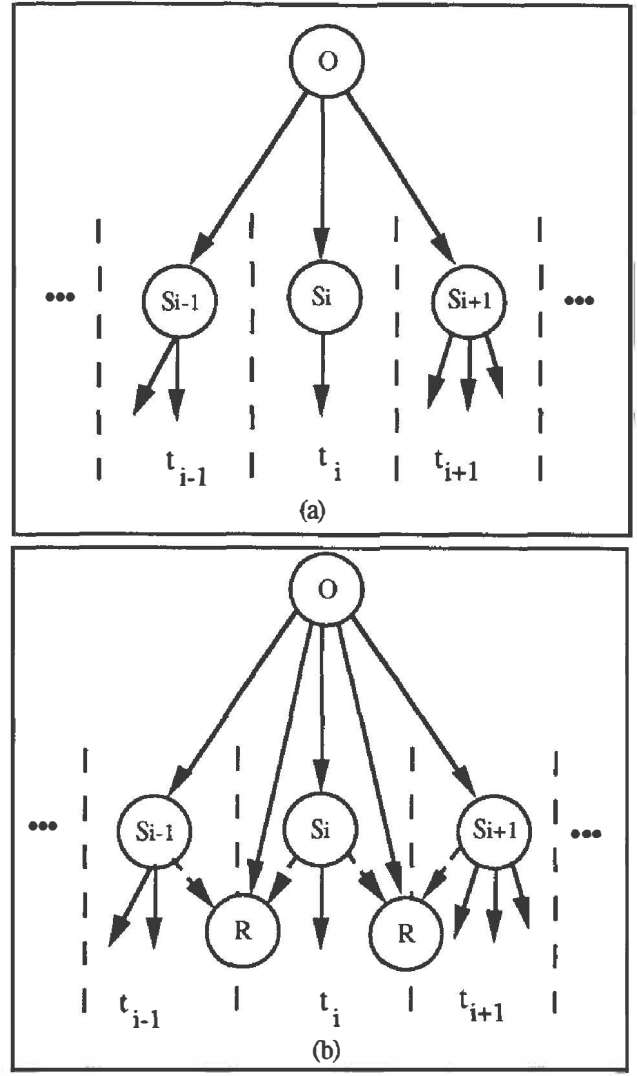

Figure 4: Dynamic recognition. The network in (a) corresponds to a probabilistic network for representing an object across several images, which is extended to include relations in (b).

## 4 APPLICATION TO COLONOSCOPY

Endoscopy is one of the tools available for diagnosis and treatment of gastrointestinal diseases. It allows a physician to obtain direct colour information of the inside surface of the human digestive system. The endoscope is a flexible tube which has a manoeuvrable tip. The orientation of the tip can be controlled by pull wires that bend the tip in 4 orthogonal directions (left/right, up/down). The consultant controls the instrument by steering the tip with two mechanical wheels, and by pushing or pulling the shaft. He can control the air supply (inflate or aspirate) for good vision but without creating excessive air pressure. He can use the water jet for cleaning the lens when it is dirty, and aspirate excess fluid. In addition to control actions he makes diagnostic decisions or therapeutic actions in each particular case. This requires a high degree of skill and experience that only an "expert" endoscopist will have. We are interested primarily in colonoscopy, which is especially difficult due to the complexity and variability of the human colon. The doctor inserts the instrument estimating the position of the colon centre (*lumen*) using several visual clues such as the darkest region, the colon muscular curves, the longitudinal muscle and others. If the tip is not controlled correctly it can be very painful and dangerous to the patient, and could even cause perforations on the colon wall. This is further complicated by the presence of many difficult situations such as the contraction and movement of the colon, fluid and bubbles that obstruct the view, pockets (*diverticula*) that can be confused with the *lumen* and the paradoxical behaviour produced by the endoscope looping inside the colon.

We are developing an expert system for colonoscopy to help the doctor with the navigation of the endoscope inside the colon and give advise to learning endoscopists suggesting correct actions. A knowledge-base in colon endoscopy was compiled with the help of an expert colonoscopist. This initial KB consisted of heuristic rules used by an endoscopist to infer a model of the colon from the visual information (interpretation) and determine the corresponding actions (control). As an initial representation we selected informal "IF-THEN" rules because they are closer to the endoscopist conceptualization's. Having extracted about 60 rules [Sucar 90], we then transformed the colonoscopy KB to a probabilistic network representation.

The qualitative knowledge-base for colonoscopy contains several rules that express relational and dynamic knowledge. To incorporate this knowledge in our probabilistic network we need to represent it within this framework using the modelling techniques developed in the previous sections. To illustrate this process we will use two rules that express relations and two more rules that incorporate dynamic information, presenting a possible representation as a probabilistic network for each one of them. Two typical examples of colonoscopy rules that involve relations are the following:

IF bright region *surrounding* a dark region

THEN diverticulum (1)

IF dark region *adjacent* to a bright arc

THEN bend in the colon (2)

These rules will be represented by the networks in fig. 5 (a) and (b), respectively.

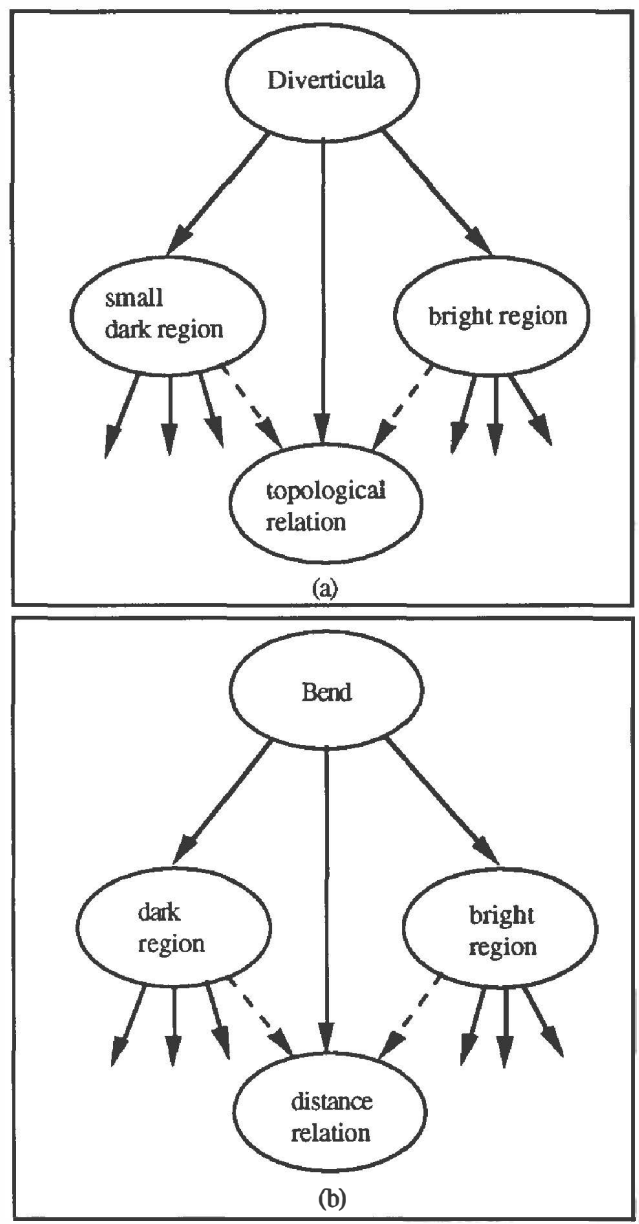

Figure 5: Examples of endoscopy relational rules. The probabilistic network in (a) represents the rule: *Diverticulum :- bright region surrounding a small dark region*, and the one in (b) corresponds to: *Bend :- bright region near to a dark region.*

Both express the fact that the recognition of an object (*diverticulum, bend*) depends on two different image objects and their relation. In the first case the relation is that one type of region is *surrounding* another, which we generalize to any type of topological relation as shown in the network. The second rule involves two image features which should be *adjacent*, and this relation is measured by the distance between the two regions. So the probabilistic network implies that the objects at the root are *dependant* on three variables: two image objects and one relation.

With respect to rules that involve temporal or dynamic knowledge we also include two examples, one from what we have defined as semi-static and the other from dynamic knowledge. The first example does not involve any particular rule, but it tries to represent the more general fact that as the endoscope moves slowly across the colon the presence or absence of the *lumen* in the image will tend to be constant across several contiguous images. That is, if we recognize a region as *lumen* at time $t_i$, there is a *high* probability that there is also a *lumen* at $t_{i+1}$ ($t_{i+1} = t_i + \Delta t$) and a *low* probability that there is *NOT lumen*. This is represented graphically by the probabilistic network in fig. 6 (a), which we simplify for recognition at time $t_i$ by the network in fig. 6 (b).

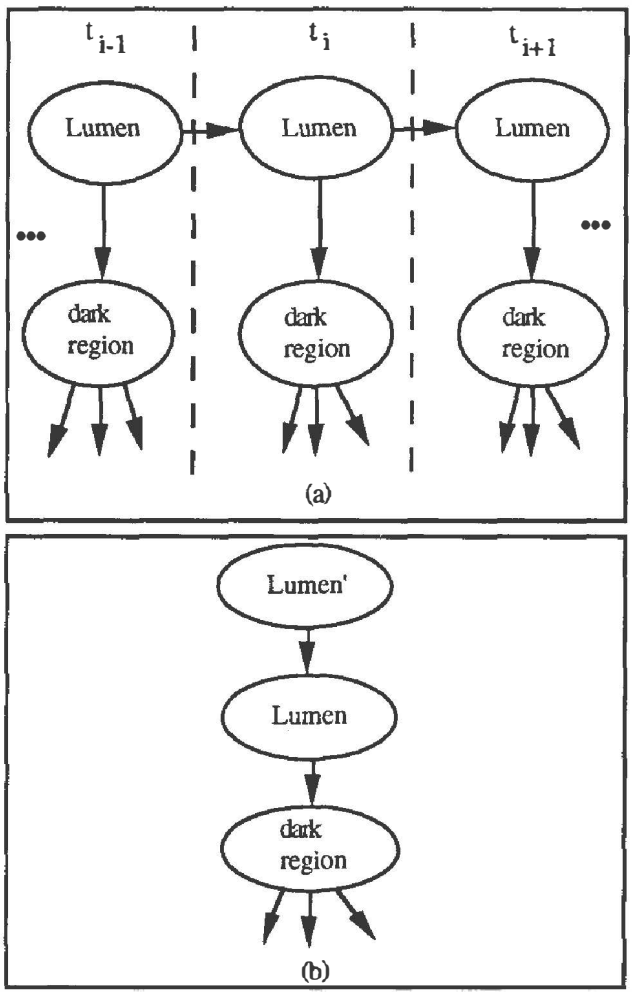

Figure 6: Use of dynamic information for *lumen* recognition. The network in (a) represents the causal relations across several successive images. This network is approximated by the one in (b) for *lumen* recognition at a given time.

If we compare the structure in figure 6 (b) with the previous tree for *lumen* recognition without using information from previous images [Sucar 91], the main difference is that the "prior" probability of *lumen* ($L$) becomes a combination of $P(L)$ and $P(L|L')$. In this case the posterior probability of *lumen* will be given by:

$$P(L|L',E) = \alpha \ [ \ P(L') \ P(L|L') \ ] \ [ \ P(E|L) \ ] \qquad (3)$$

where $E$ stands for all the current image evidence (subtree below $L$), $P(L') = P(L)$ is the *a-priori* probability of observing a *lumen*, and $\alpha$ is a normalizing constant. The first term between brackets in equation 3 corresponds to the *causal* support for $L$, in which the actual prior is combined with the results from the previous image; and the second term represents the diagnostic support which depends on the evidence obtained from the current image.

The second case of temporal reasoning is represented by the following rule:

IF yellow or green or brown spots & *static* in image
THEN lens is dirty                                         (4)

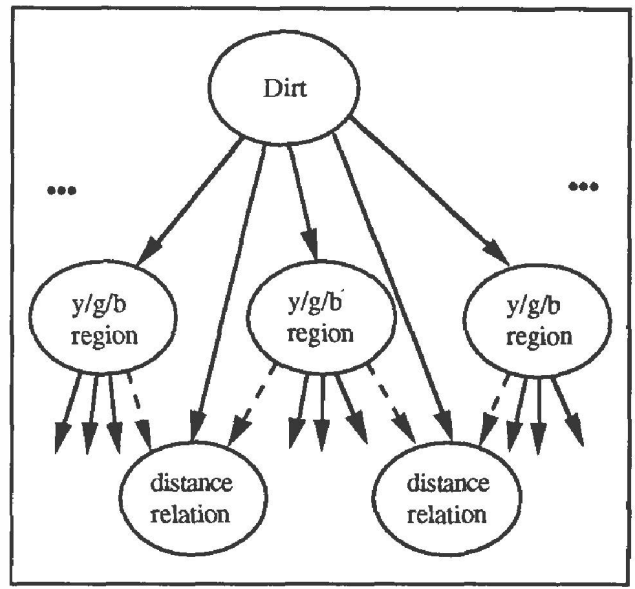

Figure 7: Probabilistic network for a *dirt spot* recognition. The yellow/green/blue (y/g/b) region nodes correspond to the same region in different successive images, and the distance relation node to their distance between two consecutive images.

This rule expresses that to identify a certain object we require that an image object is present in a succession of images, and it also specifies that certain relation must hold between the image objects. We represent this rule by the network in fig. 7, representing the concept of *static* as a zero (or close to zero) distance between the regions from several images. Although we represent the network involving an arbitrary number of images, in practice a limited number of images will be taken into consideration for recognition. In both examples involving multiple images an important assumption is that the image objects from the different successive images should *match*, having similar shape, size, colour, etc.

## 5 CONCLUSIONS

We have shown that we can represent some complex forms of knowledge using a probabilistic network, including relational and temporal information which is important for high-level vision. These extended models have a simple structure which permits us to propagate probability efficiently. They have been applied to the domain of colonoscopy, illustrating how the general modelling principles can be used in specific cases. The next stage in our research will be to test these models experimentally, obtaining the required probability distributions, validating the dependency structures and testing their performance.

### Acknowledgments

The authors will like to express their thanks to: *CONACYT* and the *Instituto de Investigaciones Electricas*, Mexico for supporting L. E. Sucar, The Olympus Optical Company, Tokyo, for providing equipment and Dr. Cristopher Williams at St. Mark's Hospital, London, for providing the expert advice.